\documentclass[sigconf]{acmart}
\usepackage{multirow}

\usepackage{amsmath}
\usepackage{multirow}
\usepackage{algorithmic}
\usepackage{amsmath}
\usepackage{CJK}
\usepackage{amsfonts}
\usepackage{bm}
\usepackage{graphicx}
\usepackage{subfigure}
\usepackage{soul}
\usepackage{amsmath}
\usepackage[utf8]{inputenc}
\usepackage{algorithm}
\usepackage{array}
\AtBeginDocument{%
  \providecommand\BibTeX{{%
    \normalfont B\kern-0.5em{\scshape i\kern-0.25em b}\kern-0.8em\TeX}}}

\setcopyright{iw3c2w3}
\copyrightyear{2021}
\acmYear{2021} 
\setcopyright{iw3c2w3}
\acmConference[WWW '21 Companion]{Companion Proceedings of the Web Conference 2021}{April 19--23, 2021}{Ljubljana, Slovenia} 
\acmBooktitle{Companion Proceedings of the Web Conference 2021 (WWW '21 Companion), April 19--23, 2021, Ljubljana, Slovenia}
\acmPrice{}
\acmDOI{10.1145/3442442.3452328}
\acmISBN{978-1-4503-8313-4/21/04}
\begin{document}

\title{How does Truth Evolve into Fake News? An Empirical Study of Fake News Evolution}

% \affiliation{%
% \author{Mingfei Guo}
% \email{1800012773@pku.edu.cn}
% % \orcid{1234-5678-9012}
% % \author{G.K.M. Tobin}
% \affiliation{%
% Wangxuan Institute of Computer Technology, Peking University,Beijing,China

\author{Mingfei Guo}
\affiliation{%
  \institution{Wangxuan Institute of Computer Technology, Peking University}
  \city{Beijing}
  \country{China}}
\email{1800012773@pku.edu.cn}
\author{Xiuying Chen}
\affiliation{%
  \institution{Wangxuan Institute of Computer Technology, Peking University}
  \city{Beijing}
  \country{China}}
\email{xy-chen@pku.edu.cn}

\author{Juntao Li}
\affiliation{%
  \institution{Soochow University}
  \city{Suzhou}
  \country{China}}
\email{ljt@suda.edu.cn}

\author{Dongyan Zhao}
\affiliation{%
  \institution{Wangxuan Institute of Computer Technology, Peking University}
  \city{Beijing}
  \country{China}}
\email{zhaody@pku.edu.cn}
\author{Rui Yan}
\affiliation{%
  \institution{Gaoling School of Artificial Intelligence, Renmin University of China}
  \city{Beijing}
  \country{China}}
\email{ruiyan@ruc.edu.cn}

\begin{abstract}
Automatically identifying fake news from the Internet is a challenging problem in deception detection tasks.
Online news is modified constantly during its propagation, e.g., malicious users distort the original truth and make up fake news.
However, the continuous evolution process would generate unprecedented fake news and cheat the original model.
We present the Fake News Evolution (FNE) dataset: a new dataset tracking the fake news evolution process.
Our dataset is composed of 950 paired data, each of which consists of articles representing the three significant phases of the evolution process, which are the truth, the fake news, and the evolved fake news.
We observe the features during the evolution and they are the disinformation techniques, text similarity, top 10 keywords, classification accuracy, parts of speech, and sentiment properties.
\end{abstract}






% \keywords{datasets, neural networks, gaze detection, text tagging}

\maketitle

\section{Introduction}
Fake news is a form of news consisting of deliberate disinformation or hoaxes according to its definition on Wikipedia, and it generally spreads via traditional news media or online social media platforms.
Sharing fake news on social media platforms is a global concern and the motivation behind it is complicated and is still a fundamental topic in psychology and sociology~\cite{llport1947ThePO}.
Usually, fake news is written and published with the intent to mislead the public to damage an agency, entity, or person, or to gain financially or politically~\cite{Wheaton2019BirthersHS,Tandoc2018DefiningN}.
It often uses sensationalist, dishonest, or outright fabricated headlines to increase readership and make a profit.
A large amount of fake news has been observed to spread online like virus outbreaks, often with alarming consequences in the real world~\cite{Carvalho2009ThePE}. 
A series of prior works~\cite{Yang2012AutomaticDO,Shu2017FakeND,Castillo2011InformationCO,Ma2015DetectRU,Hamidian2019RumorDA,Kwon2017RumorDO,Zellers2019DefendingAN,Tee2018TrustNB,Ibrishimova2019AML,Mihalcea2009TheLD} focus on automatically identifying fake news.
% and constructing classification models using various techniques.
However, in real social networks, the news evolves constantly in the spreading process, which grows more concise, more attractive, more easily grasped and told~\cite{Zhang2011RumorEI,Adamic2016InformationEI}.
This evolution process originates from the cumulative modifications during the spreading process, which is similar to a Children's game called 'Chinese Whispers' or 'Telephone'~\cite{Blackmore1999TheMM}.

%第二段：介绍数据集
% large 删掉 形容词注意
% cite 都写在一个括号里面
% Various work on the fake news spreading process has been proposed.
% In real social networks, the rumor spreading process is much more complicated than the reported scenar- ios [17–21]. The rumor evolves constantly in its spreading process, which grows shorter, more concise, more easily grasped and told [1]. The behavior originates from the cumulative modifications during the spreading process, which is called ’Chinese Whispers’ or ’Telephone’ in some conditions [22]. The phenomenon is easy to be observed in real social networks. For example, on a microblogging site Twitter, once a user discusses a certain topic, his or her followers will understand the topic indirectly. If it is a rumor, the following discussions can roughly be clas- sified as: affirmative, negative, curious, unrelated and unknown arguments [23]. 

\begin{table}[ht]
\resizebox{0.45\textwidth}{!}{
\begin{tabular}{|l|}
\hline
\begin{tabular}[c]{@{}l@{}}\emph{\textbf{Truth:}} The \hl{laws restrict people from various military-like activities} \\ that result in violence and civil disorder. The amendment to the \\ Virginia law, specifically, prohibits marching with weapons or \\ explosives for the purpose of intimidation.\end{tabular}                \\ \hline
\begin{tabular}[c]{@{}l@{}}\emph{\textbf{Fake News:}} Proposed virginia \sethlcolor{pink}\hl{law would outlaw krav maga, brazilian} \\ \sethlcolor{pink}\hl{jiu jitsu, kickboxing, tai chi, firearms instruction and self-defense} \\ training, the story reads: The law would instantly transform all martial \\ arts instructors into criminal felons.\end{tabular}         \\ \hline
\begin{tabular}[c]{@{}l@{}}\emph{\textbf{Evolved Fake News:}} The state of virginia, now entirely run by \sethlcolor{pink}\hl{truly} \\\sethlcolor{pink}\hl{insane democrats who support infanticide and child murder}, is \\proposing a new 2020 law known as sb64 (see link here) which will \\be taken up by the democrat-run senate beginning january 8, 2020.\end{tabular} \\ \hline
\end{tabular}}
\caption{An example of the dataset. We highlight the original truth information by using yellow and the fake news and the evolved fake news by using pink.}
\label{fig:intro-case}
\end{table}

We have seen that numerous recognition models have high overall performance in identifying the authenticity of news articles, and we believe these state-of-the-art models would obtain better results if they learn the features of the evolution process.
As the continuous evolution process would generate unprecedented fake news and cheat the original model, it is important to teach the model to identify the variations of fake news.
The dataset for deception detection tasks is not new to the natural language processing community, and most of them centre on labeled fake news based on human annotation or machine learning assessments~\cite{Augenstein2019MultiFCAR,Wang2017LiarLP,Shu2020FakeNewsNetAD}.
Based on these data, most efforts focuses on the identification of fake news, offering little understanding of the evolutionary process of false information.
Seldom work focus on the attribute of observing the evolution from facts to fake news to our best knowledge~\cite{Shao2016HoaxyAP}.

We introduce the Fake News Evolution (FNE) dataset to fill this margin. 
Our dataset includes 950 pieces of data, each of them contains three articles representing the three phases of the evolution process, and they are the truth, the fake news and the evolved fake news.
These three articles are derived from the same particular event and Table \ref{fig:intro-case} provides an example. Since space is limited, we only represent a short paragraph here.
Through a comprehensive analysis, we investigate the features during the evolution process, including disinformation techniques, text similarity, top 10 keywords, classification accuracy, parts of speech, and sentiment properties.
By observing the proportion of commonly used disinformation technologies, we conclude the deeper pattern of constructing fake news.
When testing classification models, evolved fake news is more easily spotted than fake news.
During the evolution process, the top keywords and the parts of speech keep consistent, while the text similarity and the sentiment properties change.

The main contributions of our work can be concluded as follows:

$\bullet$ We propose a dataset that tracks the fake news evolution process.
Each piece of data includes three articles, the truth, the fake news, and the evolved fake news, and all of them are derived from the same particular event.

% 根据数据得到哪些结论
$\bullet$ We look into the data properties to study the features during the evolution process.
Here, we observe the statistics of the disinformation techniques, text similarity, top 10 keywords, classification accuracy, parts of speech of words, and sentiment properties.

$\bullet$ Our FNE dataset can be used for semi-supervised low resources tasks and can help to improve the fake news classification models.
\section{Dataset Construction}
Our dataset aims to capture articles and pinpoint how they were modified during the dissemination and evolution process.
% The information undergoes an evolutionary process that exhibits several regularities.
% We are able to gather precisely such data, consisting of evolution traces.
% of thousands of memes, collectively comprising over 460 million individual instances propagated via Facebook. 
In online environments, the news is represented in various forms, e.g. texts, photographs, and videos. 
However, modifying the news in photograph or video forms is not readily available and is inconvenient for internet surfers.
Therefore, we focus our attention on textual news to observe how information evolves when people can simply modify it and add subjective content to it. 

In order to observe the evolution process concisely, we focus on three important phases in the process, the truth, the fake news and the evolved fake news.
The truth, which is examined by multiple sides and organizations, is guaranteed to be real and credible.
Then fake news is created by malicious users or bots after modifying the original truth and applying various counterfeit methods.
When fake news is published on online platforms, they begin to spread and evolve.
After reading the fake news, some website editors will write reports based on them and make up the evolved fake news.
These editors, no matter if they are unconscious or on purpose, are deceivers who expose misleading articles to more people.
% some gullible users will take it at face value, add comment on it and forward it to more people.
% They are unconscious and innocent deceivers and make up the evolved news. of textual news 

It is difficult to track the original true information online because social media dialogues often neglect standard protocols for reporting the information source, and it is hard to determine which media content is closest to the origin~\cite{Jang2018ACA}.
Besides, making sure that the articles of these three phases are derived from one particular event or focusing on the same topic is also a tough problem.

To tackle these two problems, we build our fake news evolution dataset by performing the following procedure.
We mainly focus on the website Snopes\footnote{https://www.snopes.com/fact-check/}, which covers a wide range of news focusing on politics, science, health and etc.
It provides fact-checking content and webpage links of fake news, which is convenient for us to gather articles of different phases during the evolution process.
First, we can easily crawl this website and collect true articles because it is transparent about examination sources and methods.
The fact-checking article will provide a summary of these fact sources and we regard it as original truth.
Then we collect fake news from the citation of the content.
Since social media platforms are increasingly more likely to regulate questionable content, it is more preferable to gather the evolved fake news on an archive website which helps you to create a copy of the webpage and store it even if the source page is removed.
Snopes includes links to evolved fake news of a snapshot website Archive Today\footnote{http://archive.today/}.
We regard the records on the archive website as evolved fake news.
Therefore, we construct a dataset tracking the information evolution process, from truth to fake news and to evolved fake news.

Table \ref{fig:intro-case} provides an example of our FNE dataset and illustrates the evolution process.
The truth says that laws are restricting military-like activities, while the fake news distorts the original facts and says the proposed laws are outlawing martial arts.
Then, except for the misrepresented laws, the evolved fake news also mentions an endorsement of democrats which is never mentioned before.
Our FNE dataset has 2 million tokens in total and covers a wide range of topics such as politics, religion, health, and economics.
% We calculate the average number of tokens (Avg. tokens) and the maximum number of tokens (Max. tokens) for the articles in three different phases, and show them in Table \ref{fig:comp_baslines}.

\section{Data Properties}

We use this dataset to perform experiments and observe the features during the evolution process.
In different stages, they include disinformation techniques, text similarity, top 10 keywords, classification accuracy, part of speech of words and sentiment properties. 

\subsection{Disinformation Techniques}

\begin{figure}[ht]
\centering
\includegraphics[scale=0.20]{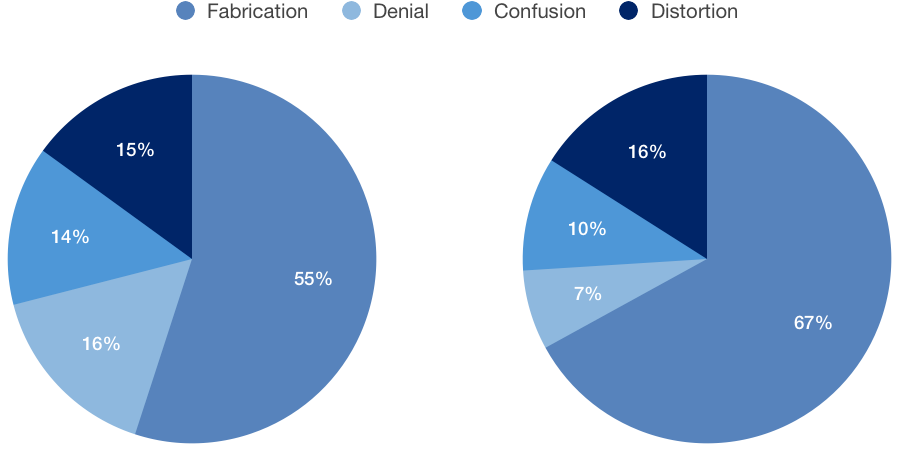}
\caption{The left one represents the proportion of disinformation techniques from truth to fake news, while the right one represents that from fake news to evolved fake news.}
\label{fig:label1}
\end{figure}

Disinformation techniques are broader and deeper patterns about how counterfeit messages are constructed~\cite{Innes2020TechniquesOD}.
To observe the popularity of various disinformation techniques in the evolution process, we ask human experts to annotate the data, that is, divide it into 4 categories, including Fabrication, Denial, Confusion, and Distortion~\cite{Martino2019FineGrainedAO}.
The kappa statistics are 0.5276 and 0.3668 for the annotated disinformation techniques from truth to fake news and from fake news to evolved fake news respectively, and that indicates the moderate agreement between annotators.

$\bullet$ \textbf{Fabrication.} 
When there exists some particular information in the fake news but not in the truth, we will suppose that this article applies Fabrication, e.g., the fake news mentions a fabricated statement in a president interview to support its political attitude.

$\bullet$ \textbf{Denial.}
This kind of fake news is generated by denying reality to avoid a psychologically uncomfortable truth~\cite{Hansson2017ScienceDA}, we can take the terms Holocaust denialism and AIDS denialism as examples.

$\bullet$ \textbf{Confusion.}
When the fake news confuses the original event with other events deliberately, we regard it as Confusion. This is generally used between two events which happen in different times and places, for instance, replacing a small accident as a devastating forest fire happens years ago to cause public panic.

$\bullet$ \textbf{Distortion.} 
Distortion means maliciously misinterpreting certain aspects of the truth, especially particular words, statistical numbers, and severe degrees.

Take Table \ref{fig:intro-case} as an example, it uses Distortion to make up fake news since it misapprehends the meaning of "military-like activities" and replaces it with "self-defense training". 
Then from the fake news to the evolved fake news, it uses Fabrication because it trumps up a democrat's endorsement.
As can be seen from Figure \ref{fig:label1}, for the frequency of disinformation techniques, the proportion of Fabrication is relatively larger, while Denial, Confusion, and Distortion are smaller. 
Since Denial, Confusion, and Distortion are largely based on prior events or well-known facts rather than fabricated information, these kinds of fake news are easily spotted.

\subsection{Text Similarity}

We calculate the similarity score between the truth and the fake news $X_1$, and that between the fake news and the evolved fake news $X_2$.
As the truth, the fake news and the evolved fake news are one-to-one correspondence, the length of $X_1$ equals that of $X_2$, $n_1=n_2$.
To assess the similarity score, We apply Word Mover’s Distance (WMD)~\cite{Kusner2015FromWE}, which enables us to evaluate the distance between two documents in a meaningful way even when they have no words in common.
It uses word2vec vector embeddings of words~\cite{Mikolov2013EfficientEO}, by matching the relevant words, WMD is able to accurately measure the similarity between two sentences.

Next, we use the t-test to determine if these two sets of scores are significantly different from each other.
We hypothesize that the variances of $X_1$ and $X_2$ populations are equal.
Here, $\overline{X_1}$ and $\overline{X_2}$ are the average scores of $X_1$ and $X_2$, while $s_{X_1}^2$ and $s_{X_2}^2$ are the unbiased estimators of the variances of the two samples.
The null hypothesis here is $\mu_1\geqslant\mu_2$ which can only be rejected.
When testing the null hypothesis, the population means of $X_1$ and $X_2$ are $\mu_1$ and $\mu_2$.
% 要写大于等于
\begin{equation}
    s_p=\sqrt{\frac{(n_1-1)*s_{X_1}^2+(n_2-1)*s_{X_2}^2}{n_1+n_2-2}}
\end{equation}
\begin{equation}
    t=\frac{\overline{X_1}-\overline{X_2}}{s_p*\sqrt{\frac{1}{n_1}+\frac{1}{n_2}}}
\end{equation}
%191
Note that $s_p$ is an estimator of the pooled standard deviation of the two samples. For significance testing, the degree of freedom for this test is $n_1+n_2-2$. We determine the $t$ value through this formula, and we find the corresponding p-value, which is the evidence against a null hypothesis, using a table of values from Student's t-distribution. We find that $t=-$39.3309.
For $\mu_1<\mu_2$, the estimated p-value is greater than 0.99, so we can reject the null hypothesis with a high degree of confidence that the similarity degree between the truth and the fake news is smaller than that between the fake news and the evolved fake news. Compared with the change between fake news and evolved fake news, the transformation between truth and fake news is more drastic and extensive.

\subsection{Keywords}
\begin{table}[ht]
\resizebox{.450\textwidth}{!}{
\begin{tabular}{|l|l|l|l|l|l|}
\hline
\multirow{2}{*}{truth}        & trump   & said      & news    & president & people    \\ \cline{2-6} 
                              & law     & one       & state   & states    & also      \\ \hline
\multirow{2}{*}{fake news}    & trump   & president & clinton & obama     & school    \\ \cline{2-6} 
                              & police  & law       & state   & bill      & health    \\ \hline
\multirow{2}{*}{evolved fake news} & trump   & obama     & people  & said      & president \\ \cline{2-6} 
                              & clinton & ago       & like    & new       & one       \\ \hline
\end{tabular}}

\caption{The top 10 keywords of the truth, the fake news and the evolved fake news.}
\label{fig:label12}
\end{table}
We extract the keywords~\cite{Matsuo2003KeywordEF} to observe the changes of information.
% \cite{Ramos2003UsingTT}
We apply a numerical statistic method, term frequency–inverse document frequency (TFIDF)~\cite{Jones1972ASI}, to reflect the importance of words in the corpus.
In each phase, we collect all the documents as a corpus and we generate a TFIDF matrix for it, then iterate over the resulting vectors to extract top keywords.
The top 10 keywords are shown in table \ref{fig:label12}, and they are basically on the same topic in different phases.
Therefore, we can conclude that in the FNE dataset, the articles during different evolution phases are largely focusing on the same subject.
Since the top 10 keywords include presidents' names and politics-related nouns, we assume that the majority of data in our dataset focuses on political topics.

%关键词
\subsection{Classification Accuracy}
\begin{table}[ht]
\resizebox{0.45\textwidth}{!}{
\begin{tabular}{|l|l|l|l|}
\hline
         & Truth  & Fake News & Evolved Fake News \\ \hline
Naive Bayes       & 0.2084 & 0.8589    & 0.8432            \\ \hline
Random Forest & 0.4684 & 0.6274 & 0.5853 \\ \hline
Adaptive Boosting & 0.7126 & 0.1874    & 0.1579            \\ \hline
% GBDT     & 0.6716 & 0.4895    & 0.5316            \\ \hline
\end{tabular}}
\caption{Classification accuracy by using different methods.}
\label{fig:kkk}
\end{table}
We train several standard machine learning models like Naive Bayes~\cite{manning2008introduction}, Random Forest~\cite{Breiman2004RandomF}, and Adaptive Boosting~\cite{Freund1995ADG} on the LIAR dataset~\cite{Wang2017LiarLP}.
The LIAR dataset includes manually labeled statements from political websites, and its topic is consistent with our FNE dataset.
Then, we show the classification accuracy of these trained models on our FNE dataset in Table \ref{fig:kkk}.
% \cite{zellers2019grover}.
The accuracy of the evolved fake news is smaller than that of the fake news, which indicates that detecting the evolved fake news is more intractable.

\subsection{Part of Speech}
%用词类型
\begin{figure}[ht]
\centering
\includegraphics[scale=0.275]{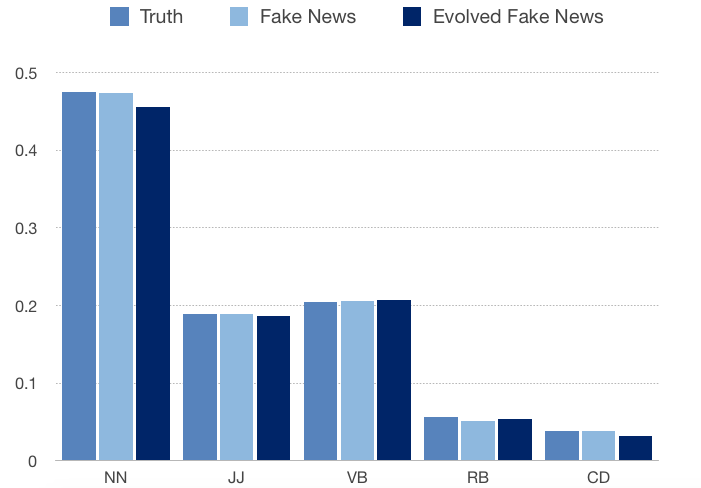}
\caption{Proportion of different parts of speech of words during the evolution process. }
\label{fig:label11}
\end{figure}
We observe the proportion of various parts of speech of the articles in three phases by using Natural Language Toolkit (NLTK)~\cite{Bird2008MultidisciplinaryIW}.
There are numerous parts of speech in the English language, and we only focus on those who take up a large proportion.
Here, we count 5 parts, and they are nouns, adjectives, verbs, adverbs, and numbers.
The result is shown in Figure \ref{fig:label11}, NN represents nouns, JJ represents adjectives, VB represents verbs, RB represents adverbs, and CD represents cardinal numbers.
NN includes singular nouns, plural nouns, and proper nouns and VB includes 3rd person verbs, non-3rd person verbs and all tenses of verbs.
% past tense

The proportion of parts of speech of the truth, the fake news, and the evolved fake news are substantially consistent.
Therefore, the authenticity of the articles only makes a small contribution to grammatical features.
However, we notice that in the fake news and the evolved fake news, the behavior of mentioning other users or media (through $@$ or $\#$) increases. The total frequency of tagging others is 368 in the truth, 696 in the fake news, and 619 in the evolved fake news.
Due to human nature, people are vulnerable to being deceived by more convincing and inflammatory news, which can be reflected by the act of citing others' example or exposition~\cite{Castillo2011InformationCO}.
% Since the human writer's personal habits contribute to the grammatical features.
% rather than the authenticity of the information.
% In the process of news evolution, the text genre remains unchanged, thus the word characteristic is correspondent.
% However, we discovered that as the deception level of fake news increases, in the evolved fake news, the behavior of tagging other users also increases~\cite{Castillo2011InformationCO}.
% Due to the intention of fake news and the human nature, people are vulnerable to be deceived by more inflammatory news, which can be reflected by the act of tagging others. 
% The more deceptive a hoax is, the more likely for people to be cheated, and the easier for it to spread further. This result is in line with human intuition.

\subsection{Sentiment}
% \begin{figure}[ht]
% \centering
% \includegraphics[scale=0.42]{bar.png}
% \caption{Sentiment and Subjective/Objective Score in Evolution Process.}
% \label{fig:label3}
% \end{figure}
%算情感分析的分数和主客观的分数，说明信息是怎样被改变的
% vice versa
% 语法错误 ~\cite{Pang2007OpinionMA}
We calculate the sentiment property of the articles using TextBlob to observe the tendency during the evolution process.
The sentiment property includes two scores, polarity and subjectivity.
For each text, the value of polarity is in the range of -1 to 1. The value of -1 means very negative, and the score of 1 means very positive. Similarly, the value of subjectivity is in the range of 0 to 1, where 0 is very objective and 1 is very subjective.

To analyze the differences among the sentiment properties among the truth, the fake news and the evolved fake news, we use analysis of variance (ANOVA), which can decide whether the sample means are equal or not.
The p-value corresponding to the F-statistic of ANOVA is lower than 0.01 (equals to 99\% confidence coefficient), suggesting that one or more treatments are significantly different.

Then, we use Tukey's test, which applies simultaneously to the set of all pairwise comparisons and can find means that are significantly different from each other.
We first establish the critical value $Q$ statistic of the Tukey's test based on $k=$ 3 samples and $v=$ 2847 degrees of freedom.
For significance level 99\% ($\alpha=$ 0.01) in the Studentized Range distribution, $Q_{critical}^{\alpha=0.01,k=3,\nu=2847}=$ 4.1237. Then we calculate a parameter for each pair of samples being compared:
\begin{equation}
    Q_{i, \,j} = \frac{\lvert
\bar{x}_{i} - \bar{x}_{j} \rvert }{s_{i, \,j}}
\end{equation}
\begin{equation}
\,\, {s_{i, \,j}} = \frac{\hat{\sigma}_\epsilon
}{\sqrt{H_{i, \, j} } } \,\,\,\,\, i, \, j = 1, \, \ldots, \, k; \,\, i \ne j
\, .
\end{equation}
The quantity ${H_{i, \, j}}$ is the harmonic mean of the number of observations in samples labeled $i$ and $j$, and ${H_{i, \, j}}=$ 950.
$\hat{\sigma}_\epsilon$ is the square root of the mean square error determined in the precursor ANOVA procedure, and it is same across all pairs being compared.
We evaluate whether $Q_{i, \,j} > Q_{critical}$ to decide if we can reject the null hypothesis. Here, reject the null hypothesis means that a significant difference has been observed. 
The result is shown in Table \ref{fig:label3}.

\begin{table}[ht]
\resizebox{.450\textwidth}{!}{
\begin{tabular}{|l|l|l|l|}
\hline
             & Truth vs Fake & Truth vs Evolved & Fake vs Evolved \\ \hline
Polarity     & 0.1735             & 7.2598                     & 7.0863                         \\ \hline
Subjectivity &   0.2696                 &     6.8796                       &               7.1492                 \\ \hline
\end{tabular}}
\caption{$Q$ statistic of the Tukey's test among the three samples, the truth, the fake news and the evolved fake news.}
\label{fig:label3}
\end{table}

For both the polarity and the subjectivity scores, the evolved fake news is the one that is significantly different from others.
We calculate the average scores, the polarity of the truth, the fake and the evolved fake is 0.0357, 0.0361, and 0.0523, while the subjectivity is 0.3474, 0.3482, and 0.3263.
Generally, the articles of this dataset are objective and positive.
We use the t-test mentioned in the previous section again and discover that we can say with 99\% confidence that the polarity score of the evolved fake news is larger and the subjectivity score is smaller than that of the truth and the fake news.
Therefore, the evolved fake news is more positive and more objective than the truth and the fake news.
We assume that the evolved fake news is designed to be formal and convincing, therefore it can mislead people effectively~\cite{Guo2019ExploitingEF}, and this is consistent with prior classification results.
% The truth and the fake news 

\section{Conclusion}
%总结
We introduce a new fake news evolution dataset.
Compared with prior datasets, our FNE dataset reflects the evolution process of fake news on the Internet. 
Each piece of data is composed of the truth, the fake news, and the evolved fake news, and all three articles are derived from the same event.
% We look into the data properties and draw some conclusions.
During the evolution process, the disinformation techniques, parts of speech, and keywords keep consistent, while the text similarity and sentiment change.
The similarity score between the fake news and the evolved fake news is larger than that between the truth and the fake news.
For sentiment properties, the evolved fake news is more objective and positive, and it is more difficult to be detected.
In future work, we would like to use this dataset on fake news recognition models to promote classification performance when facing newly generated rumors. 
We hope the release of this dataset could contribute to the development of fake news detection and accelerate further investigation into the evolution process.
% solving the problems that need to track the evolution process.

% \section{Citations and Bibliographies}
\bibliographystyle{ACM-Reference-Format}
\bibliography{main}


\begin{thebibliography}{34}

%%% the Web address.

\ifx \showCODEN    \undefined \def \showCODEN     #1{\unskip}     \fi
\ifx \showDOI      \undefined \def \showDOI       #1{#1}\fi
\ifx \showISBNx    \undefined \def \showISBNx     #1{\unskip}     \fi
\ifx \showISBNxiii \undefined \def \showISBNxiii  #1{\unskip}     \fi
\ifx \showISSN     \undefined \def \showISSN      #1{\unskip}     \fi
\ifx \showLCCN     \undefined \def \showLCCN      #1{\unskip}     \fi
\ifx \shownote     \undefined \def \shownote      #1{#1}          \fi
\ifx \showarticletitle \undefined \def \showarticletitle #1{#1}   \fi
\ifx \showURL      \undefined \def \showURL       {\relax}        \fi
% invisible to TeX
\providecommand\bibfield[2]{#2}
\providecommand\bibinfo[2]{#2}
\providecommand\natexlab[1]{#1}
\providecommand\showeprint[2][]{arXiv:#2}

\bibitem[\protect\citeauthoryear{Adamic, Lento, Adar, and Ng}{Adamic
  et~al\mbox{.}}{2016}]%
        {Adamic2016InformationEI}
\bibfield{author}{\bibinfo{person}{Lada~A. Adamic}, \bibinfo{person}{Thomas~M.
  Lento}, \bibinfo{person}{Eytan Adar}, {and} \bibinfo{person}{Pauline~C. Ng}.}
  \bibinfo{year}{2016}\natexlab{}.
\newblock \showarticletitle{Information Evolution in Social Networks}. In
  \bibinfo{booktitle}{\emph{WSDM '16}}.
\newblock


\bibitem[\protect\citeauthoryear{Augenstein, Lioma, Wang, Lima, Hansen, Hansen,
  and Simonsen}{Augenstein et~al\mbox{.}}{2019}]%
        {Augenstein2019MultiFCAR}
\bibfield{author}{\bibinfo{person}{Isabelle Augenstein},
  \bibinfo{person}{Christina Lioma}, \bibinfo{person}{Dongsheng Wang},
  \bibinfo{person}{Lucas~Chaves Lima}, \bibinfo{person}{Casper Hansen},
  \bibinfo{person}{Christian Hansen}, {and} \bibinfo{person}{Jakob~Grue
  Simonsen}.} \bibinfo{year}{2019}\natexlab{}.
\newblock \showarticletitle{MultiFC: A Real-World Multi-Domain Dataset for
  Evidence-Based Fact Checking of Claims}. In
  \bibinfo{booktitle}{\emph{EMNLP/IJCNLP}}.
\newblock


\bibitem[\protect\citeauthoryear{Bird, Klein, Loper, and Baldridge}{Bird
  et~al\mbox{.}}{2008}]%
        {Bird2008MultidisciplinaryIW}
\bibfield{author}{\bibinfo{person}{Steven Bird}, \bibinfo{person}{E. Klein},
  \bibinfo{person}{E. Loper}, {and} \bibinfo{person}{Jason Baldridge}.}
  \bibinfo{year}{2008}\natexlab{}.
\newblock \showarticletitle{Multidisciplinary Instruction with the Natural
  Language Toolkit}.
\newblock


\bibitem[\protect\citeauthoryear{Blackmore}{Blackmore}{1999}]%
        {Blackmore1999TheMM}
\bibfield{author}{\bibinfo{person}{Susan Blackmore}.}
  \bibinfo{year}{1999}\natexlab{}.
\newblock \showarticletitle{The Meme Machine}.
\newblock


\bibitem[\protect\citeauthoryear{Breiman}{Breiman}{2004}]%
        {Breiman2004RandomF}
\bibfield{author}{\bibinfo{person}{L. Breiman}.}
  \bibinfo{year}{2004}\natexlab{}.
\newblock \showarticletitle{Random Forests}.
\newblock \bibinfo{journal}{\emph{Machine Learning}}  \bibinfo{volume}{45}
  (\bibinfo{year}{2004}), \bibinfo{pages}{5--32}.
\newblock


\bibitem[\protect\citeauthoryear{Carvalho, Klagge, and Moench}{Carvalho
  et~al\mbox{.}}{2009}]%
        {Carvalho2009ThePE}
\bibfield{author}{\bibinfo{person}{C. Carvalho}, \bibinfo{person}{Nicholas
  Klagge}, {and} \bibinfo{person}{Emanuel Moench}.}
  \bibinfo{year}{2009}\natexlab{}.
\newblock \showarticletitle{The Persistent Effects of a False News Shock}.
\newblock \bibinfo{journal}{\emph{Behavioral \& Experimental Finance eJournal}}
  (\bibinfo{year}{2009}).
\newblock


\bibitem[\protect\citeauthoryear{Castillo, Mendoza, and Poblete}{Castillo
  et~al\mbox{.}}{2011}]%
        {Castillo2011InformationCO}
\bibfield{author}{\bibinfo{person}{Carlos Castillo}, \bibinfo{person}{Marcelo
  Mendoza}, {and} \bibinfo{person}{Barbara Poblete}.}
  \bibinfo{year}{2011}\natexlab{}.
\newblock \showarticletitle{Information credibility on twitter}. In
  \bibinfo{booktitle}{\emph{WWW}}.
\newblock


\bibitem[\protect\citeauthoryear{Freund and Schapire}{Freund and
  Schapire}{1995}]%
        {Freund1995ADG}
\bibfield{author}{\bibinfo{person}{Y. Freund} {and} \bibinfo{person}{R.
  Schapire}.} \bibinfo{year}{1995}\natexlab{}.
\newblock \showarticletitle{A decision-theoretic generalization of on-line
  learning and an application to boosting}. In
  \bibinfo{booktitle}{\emph{EuroCOLT}}.
\newblock


\bibitem[\protect\citeauthoryear{Guo, Cao, Zhang, Shu, and Yu}{Guo
  et~al\mbox{.}}{2019}]%
        {Guo2019ExploitingEF}
\bibfield{author}{\bibinfo{person}{Chuan Guo}, \bibinfo{person}{J. Cao},
  \bibinfo{person}{X. Zhang}, \bibinfo{person}{Kai Shu}, {and}
  \bibinfo{person}{M. Yu}.} \bibinfo{year}{2019}\natexlab{}.
\newblock \showarticletitle{Exploiting Emotions for Fake News Detection on
  Social Media}.
\newblock \bibinfo{journal}{\emph{ArXiv}}  \bibinfo{volume}{abs/1903.01728}
  (\bibinfo{year}{2019}).
\newblock


\bibitem[\protect\citeauthoryear{Hamidian and Diab}{Hamidian and Diab}{2019}]%
        {Hamidian2019RumorDA}
\bibfield{author}{\bibinfo{person}{Sardar Hamidian} {and}
  \bibinfo{person}{Mona~T. Diab}.} \bibinfo{year}{2019}\natexlab{}.
\newblock \showarticletitle{Rumor Detection and Classification for Twitter
  Data}.
\newblock \bibinfo{journal}{\emph{ArXiv}}  \bibinfo{volume}{abs/1912.08926}
  (\bibinfo{year}{2019}).
\newblock


\bibitem[\protect\citeauthoryear{Hansson}{Hansson}{2017}]%
        {Hansson2017ScienceDA}
\bibfield{author}{\bibinfo{person}{S. Hansson}.}
  \bibinfo{year}{2017}\natexlab{}.
\newblock \showarticletitle{Science denial as a form of pseudoscience.}
\newblock \bibinfo{journal}{\emph{Studies in history and philosophy of
  science}}  \bibinfo{volume}{63} (\bibinfo{year}{2017}),
  \bibinfo{pages}{39--47}.
\newblock


\bibitem[\protect\citeauthoryear{Ibrishimova and Li}{Ibrishimova and
  Li}{2019}]%
        {Ibrishimova2019AML}
\bibfield{author}{\bibinfo{person}{Marina~Danchovsky Ibrishimova} {and}
  \bibinfo{person}{Kin~Fun Li}.} \bibinfo{year}{2019}\natexlab{}.
\newblock \showarticletitle{A Machine Learning Approach to Fake News Detection
  Using Knowledge Verification and Natural Language Processing}. In
  \bibinfo{booktitle}{\emph{INCoS}}.
\newblock


\bibitem[\protect\citeauthoryear{Innes}{Innes}{2020}]%
        {Innes2020TechniquesOD}
\bibfield{author}{\bibinfo{person}{Martin Innes}.}
  \bibinfo{year}{2020}\natexlab{}.
\newblock \showarticletitle{Techniques of disinformation: Constructing and
  communicating “soft facts” after terrorism}.
\newblock \bibinfo{journal}{\emph{The British Journal of Sociology}}
  \bibinfo{volume}{71} (\bibinfo{year}{2020}), \bibinfo{pages}{284 -- 299}.
\newblock


\bibitem[\protect\citeauthoryear{Jang, Geng, Li, Xia, Huang, Kim, and
  Tang}{Jang et~al\mbox{.}}{2018}]%
        {Jang2018ACA}
\bibfield{author}{\bibinfo{person}{S.~Mo Jang}, \bibinfo{person}{Tieming Geng},
  \bibinfo{person}{Jo-Yun~Queenie Li}, \bibinfo{person}{Ruofan Xia},
  \bibinfo{person}{Chin-Tser Huang}, \bibinfo{person}{Hwalbin Kim}, {and}
  \bibinfo{person}{Jijun Tang}.} \bibinfo{year}{2018}\natexlab{}.
\newblock \showarticletitle{A computational approach for examining the roots
  and spreading patterns of fake news: Evolution tree analysis}.
\newblock \bibinfo{journal}{\emph{Comput. Hum. Behav.}}  \bibinfo{volume}{84}
  (\bibinfo{year}{2018}), \bibinfo{pages}{103--113}.
\newblock


\bibitem[\protect\citeauthoryear{Jones}{Jones}{1972}]%
        {Jones1972ASI}
\bibfield{author}{\bibinfo{person}{K. Jones}.} \bibinfo{year}{1972}\natexlab{}.
\newblock \showarticletitle{A statistical interpretation of term specificity
  and its application in retrieval}.
\newblock \bibinfo{journal}{\emph{Journal of Documentation}}
  \bibinfo{volume}{60} (\bibinfo{year}{1972}), \bibinfo{pages}{493--502}.
\newblock


\bibitem[\protect\citeauthoryear{Kusner, Sun, Kolkin, and Weinberger}{Kusner
  et~al\mbox{.}}{2015}]%
        {Kusner2015FromWE}
\bibfield{author}{\bibinfo{person}{Matt~J. Kusner}, \bibinfo{person}{Yu Sun},
  \bibinfo{person}{Nicholas~I. Kolkin}, {and} \bibinfo{person}{Kilian~Q.
  Weinberger}.} \bibinfo{year}{2015}\natexlab{}.
\newblock \showarticletitle{From Word Embeddings To Document Distances}. In
  \bibinfo{booktitle}{\emph{ICML}}.
\newblock


\bibitem[\protect\citeauthoryear{Kwon, Cha, and Jung}{Kwon
  et~al\mbox{.}}{2017}]%
        {Kwon2017RumorDO}
\bibfield{author}{\bibinfo{person}{Sejeong Kwon}, \bibinfo{person}{Meeyoung
  Cha}, {and} \bibinfo{person}{Kyomin Jung}.} \bibinfo{year}{2017}\natexlab{}.
\newblock \showarticletitle{Rumor Detection over Varying Time Windows}.
\newblock \bibinfo{journal}{\emph{PLoS ONE}}  \bibinfo{volume}{12}
  (\bibinfo{year}{2017}).
\newblock


\bibitem[\protect\citeauthoryear{Ma, Gao, Wei, Lu, and Wong}{Ma
  et~al\mbox{.}}{2015}]%
        {Ma2015DetectRU}
\bibfield{author}{\bibinfo{person}{Jing Ma}, \bibinfo{person}{Wei Gao},
  \bibinfo{person}{Zhongyu Wei}, \bibinfo{person}{Yueming Lu}, {and}
  \bibinfo{person}{Kam-Fai Wong}.} \bibinfo{year}{2015}\natexlab{}.
\newblock \showarticletitle{Detect Rumors Using Time Series of Social Context
  Information on Microblogging Websites}. In \bibinfo{booktitle}{\emph{CIKM
  '15}}.
\newblock


\bibitem[\protect\citeauthoryear{Manning, Raghavan, and Sch{\"u}tze}{Manning
  et~al\mbox{.}}{2008}]%
        {manning2008introduction}
\bibfield{author}{\bibinfo{person}{Christopher~D Manning},
  \bibinfo{person}{Prabhakar Raghavan}, {and} \bibinfo{person}{Hinrich
  Sch{\"u}tze}.} \bibinfo{year}{2008}\natexlab{}.
\newblock \bibinfo{booktitle}{\emph{Introduction to information retrieval}}.
\newblock \bibinfo{publisher}{Cambridge university press}.
\newblock


\bibitem[\protect\citeauthoryear{Martino, Yu, Barr{\'o}n-Cede{\~n}o, Petrov,
  and Nakov}{Martino et~al\mbox{.}}{2019}]%
        {Martino2019FineGrainedAO}
\bibfield{author}{\bibinfo{person}{Giovanni Da~San Martino},
  \bibinfo{person}{Seunghak Yu}, \bibinfo{person}{Alberto
  Barr{\'o}n-Cede{\~n}o}, \bibinfo{person}{Rostislav Petrov}, {and}
  \bibinfo{person}{Preslav Nakov}.} \bibinfo{year}{2019}\natexlab{}.
\newblock \showarticletitle{Fine-Grained Analysis of Propaganda in News
  Articles}.
\newblock \bibinfo{journal}{\emph{ArXiv}}  \bibinfo{volume}{abs/1910.02517}
  (\bibinfo{year}{2019}).
\newblock


\bibitem[\protect\citeauthoryear{Matsuo and Ishizuka}{Matsuo and
  Ishizuka}{2003}]%
        {Matsuo2003KeywordEF}
\bibfield{author}{\bibinfo{person}{Yutaka Matsuo} {and}
  \bibinfo{person}{Mitsuru Ishizuka}.} \bibinfo{year}{2003}\natexlab{}.
\newblock \showarticletitle{Keyword extraction from a single document using
  word co-occurrence statistical information}.
\newblock \bibinfo{journal}{\emph{Int. J. Artif. Intell. Tools}}
  \bibinfo{volume}{13} (\bibinfo{year}{2003}), \bibinfo{pages}{157--169}.
\newblock


\bibitem[\protect\citeauthoryear{Mihalcea and Strapparava}{Mihalcea and
  Strapparava}{2009}]%
        {Mihalcea2009TheLD}
\bibfield{author}{\bibinfo{person}{R. Mihalcea} {and} \bibinfo{person}{C.
  Strapparava}.} \bibinfo{year}{2009}\natexlab{}.
\newblock \showarticletitle{The Lie Detector: Explorations in the Automatic
  Recognition of Deceptive Language}. In
  \bibinfo{booktitle}{\emph{ACL/IJCNLP}}.
\newblock


\bibitem[\protect\citeauthoryear{Mikolov, Chen, Corrado, and Dean}{Mikolov
  et~al\mbox{.}}{2013}]%
        {Mikolov2013EfficientEO}
\bibfield{author}{\bibinfo{person}{Tomas Mikolov}, \bibinfo{person}{Kai Chen},
  \bibinfo{person}{G.~S. Corrado}, {and} \bibinfo{person}{J. Dean}.}
  \bibinfo{year}{2013}\natexlab{}.
\newblock \showarticletitle{Efficient Estimation of Word Representations in
  Vector Space}. In \bibinfo{booktitle}{\emph{ICLR}}.
\newblock


\bibitem[\protect\citeauthoryear{Shao, Ciampaglia, Flammini, and Menczer}{Shao
  et~al\mbox{.}}{2016}]%
        {Shao2016HoaxyAP}
\bibfield{author}{\bibinfo{person}{Chengcheng Shao},
  \bibinfo{person}{Giovanni~Luca Ciampaglia}, \bibinfo{person}{Alessandro
  Flammini}, {and} \bibinfo{person}{Filippo Menczer}.}
  \bibinfo{year}{2016}\natexlab{}.
\newblock \showarticletitle{Hoaxy: A Platform for Tracking Online
  Misinformation}. In \bibinfo{booktitle}{\emph{WWW}}.
\newblock


\bibitem[\protect\citeauthoryear{Shu, Mahudeswaran, Wang, Lee, and Liu}{Shu
  et~al\mbox{.}}{2020}]%
        {Shu2020FakeNewsNetAD}
\bibfield{author}{\bibinfo{person}{Kai Shu}, \bibinfo{person}{Deepak
  Mahudeswaran}, \bibinfo{person}{Suhang Wang}, \bibinfo{person}{Dongwon Lee},
  {and} \bibinfo{person}{Huan Liu}.} \bibinfo{year}{2020}\natexlab{}.
\newblock \showarticletitle{FakeNewsNet: A Data Repository with News Content,
  Social Context, and Spatiotemporal Information for Studying Fake News on
  Social Media}.
\newblock \bibinfo{journal}{\emph{Big data}}  \bibinfo{volume}{8 3}
  (\bibinfo{year}{2020}), \bibinfo{pages}{171--188}.
\newblock


\bibitem[\protect\citeauthoryear{Shu, Sliva, Wang, Tang, and Liu}{Shu
  et~al\mbox{.}}{2017}]%
        {Shu2017FakeND}
\bibfield{author}{\bibinfo{person}{Kai Shu}, \bibinfo{person}{Amy Sliva},
  \bibinfo{person}{Suhang Wang}, \bibinfo{person}{Jiliang Tang}, {and}
  \bibinfo{person}{Huan Liu}.} \bibinfo{year}{2017}\natexlab{}.
\newblock \showarticletitle{Fake News Detection on Social Media: A Data Mining
  Perspective}.
\newblock \bibinfo{journal}{\emph{ArXiv}}  \bibinfo{volume}{abs/1708.01967}
  (\bibinfo{year}{2017}).
\newblock


\bibitem[\protect\citeauthoryear{Tandoc, Lim, and Ling}{Tandoc
  et~al\mbox{.}}{2018}]%
        {Tandoc2018DefiningN}
\bibfield{author}{\bibinfo{person}{Edson~C. Tandoc}, \bibinfo{person}{Zheng~Wei
  Lim}, {and} \bibinfo{person}{R. Ling}.} \bibinfo{year}{2018}\natexlab{}.
\newblock \showarticletitle{Defining “Fake News”}.
\newblock \bibinfo{journal}{\emph{Digital Journalism}}  \bibinfo{volume}{6}
  (\bibinfo{year}{2018}), \bibinfo{pages}{137 -- 153}.
\newblock


\bibitem[\protect\citeauthoryear{Tee and Murugesan}{Tee and Murugesan}{2018}]%
        {Tee2018TrustNB}
\bibfield{author}{\bibinfo{person}{Wee~Jing Tee} {and}
  \bibinfo{person}{Raja~Kumar Murugesan}.} \bibinfo{year}{2018}\natexlab{}.
\newblock \showarticletitle{Trust Network, Blockchain and Evolution in Social
  Media to Build Trust and Prevent Fake News}.
\newblock \bibinfo{journal}{\emph{2018 Fourth International Conference on
  Advances in Computing, Communication \& Automation (ICACCA)}}
  (\bibinfo{year}{2018}), \bibinfo{pages}{1--6}.
\newblock


\bibitem[\protect\citeauthoryear{Wang}{Wang}{2017}]%
        {Wang2017LiarLP}
\bibfield{author}{\bibinfo{person}{William~Yang Wang}.}
  \bibinfo{year}{2017}\natexlab{}.
\newblock \showarticletitle{"Liar, Liar Pants on Fire": A New Benchmark Dataset
  for Fake News Detection}. In \bibinfo{booktitle}{\emph{ACL}}.
\newblock


\bibitem[\protect\citeauthoryear{Wheaton}{Wheaton}{2019}]%
        {Wheaton2019BirthersHS}
\bibfield{author}{\bibinfo{person}{Grace~Claire Wheaton}.}
  \bibinfo{year}{2019}\natexlab{}.
\newblock \showarticletitle{Birthers, Hand Signals, and Spirit Cooking: The
  Impact of Political Fake News Content on Facebook Engagement during the 2016
  Presidential Election}.
\newblock


\bibitem[\protect\citeauthoryear{Yang, Liu, Yu, and Yang}{Yang
  et~al\mbox{.}}{2012}]%
        {Yang2012AutomaticDO}
\bibfield{author}{\bibinfo{person}{Fan Yang}, \bibinfo{person}{Yang Liu},
  \bibinfo{person}{Xiaohui Yu}, {and} \bibinfo{person}{Min Yang}.}
  \bibinfo{year}{2012}\natexlab{}.
\newblock \showarticletitle{Automatic detection of rumor on Sina Weibo}. In
  \bibinfo{booktitle}{\emph{MDS '12}}.
\newblock


\bibitem[\protect\citeauthoryear{Zellers, Holtzman, Rashkin, Bisk, Farhadi,
  Roesner, and Choi}{Zellers et~al\mbox{.}}{2019}]%
        {Zellers2019DefendingAN}
\bibfield{author}{\bibinfo{person}{Rowan Zellers}, \bibinfo{person}{Ari
  Holtzman}, \bibinfo{person}{Hannah Rashkin}, \bibinfo{person}{Yonatan Bisk},
  \bibinfo{person}{Ali Farhadi}, \bibinfo{person}{Franziska Roesner}, {and}
  \bibinfo{person}{Yejin Choi}.} \bibinfo{year}{2019}\natexlab{}.
\newblock \showarticletitle{Defending Against Neural Fake News}.
\newblock  (\bibinfo{year}{2019}).
\newblock


\bibitem[\protect\citeauthoryear{Zhang, Zhou, Guan, and Zhou}{Zhang
  et~al\mbox{.}}{2011}]%
        {Zhang2011RumorEI}
\bibfield{author}{\bibinfo{person}{Yichao Zhang}, \bibinfo{person}{Shi Zhou},
  \bibinfo{person}{Jihong Guan}, {and} \bibinfo{person}{Shuigeng Zhou}.}
  \bibinfo{year}{2011}\natexlab{}.
\newblock \showarticletitle{Rumor Evolution in Social Networks}.
\newblock \bibinfo{journal}{\emph{arXiv: Physics and Society}}
  (\bibinfo{year}{2011}).
\newblock


\bibitem[\protect\citeauthoryear{Āllport and Postman}{Āllport and
  Postman}{1947}]%
        {llport1947ThePO}
\bibfield{author}{\bibinfo{person}{G. Āllport} {and} \bibinfo{person}{L.
  Postman}.} \bibinfo{year}{1947}\natexlab{}.
\newblock \showarticletitle{The psychology of rumor}.
\newblock


\end{thebibliography}
\end{document}